\documentclass[sigconf,anonymous=false]{acmart}
\usepackage{multirow}
\usepackage{amsmath,array,graphicx}
\usepackage{makecell}
\usepackage{subfigure}
\AtBeginDocument{%
	\providecommand\BibTeX{{%
			\normalfont B\kern-0.5em{\scshape i\kern-0.25em b}\kern-0.8em\TeX}}}

\setcopyright{acmcopyright}
\acmConference[CAAI '22]{CAAI International Conference on Artificial Intelligence}{27-28 August, 2022}{Beijing, China}
\acmBooktitle{CAAI '22: CAAI International Conference on Artificial Intelligence August 27th - 28th, 2022, Beijing, China}
\acmPrice{15.00}
\acmISBN{978-1-4503-XXXX-X/18/06}

\begin{document}
	\title{PHN: Parallel heterogeneous network with soft gating for CTR prediction}
	
	\author{Ri Su}
    \email{suricsu@csu.edu.cn}
	\affiliation{%
		\institution{Central South University}
		\state{Changsha}
		\country{China}}

    \author{Alphonse Houssou HOUNYE}
    \email{hounyea@csu.edu.cn}
	\affiliation{%
		\institution{Central South University}
		\state{Changsha}
		\country{China}}

    \author{Cong Cao}
    \email{congcao@csu.edu.cn}
	\affiliation{%
		\institution{Central South University}
		\state{Changsha}
		\country{China}}

    \author{Muzhou Hou}
    \authornote{Corresponding author.}
    \email{houmuzhou@sina.com}
	\affiliation{%
		\institution{Central South University}
		\state{Changsha}
		\country{China}}
	
	\renewcommand{\shortauthors}{Ri, et al.}
	
	\begin{abstract}
		The Click-though Rate (CTR) prediction task is a basic task in recommendation system. Most of the previous researches of CTR models built based on Wide \& deep structure and gradually evolved into parallel structures with different modules. However, the simple accumulation of parallel structures can lead to higher structural complexity and longer training time. Based on the Sigmoid activation function of output layer, the linear addition activation value of parallel structures in the training process is easy to make the samples fall into the weak gradient interval, resulting in the phenomenon of weak gradient, and reducing the effectiveness of training. To this end, this paper proposes a Parallel Heterogeneous Network (PHN) model, which constructs a network with parallel structure through three different interaction analysis methods, and uses Soft Selection Gating (SSG) to feature heterogeneous data with different structure. Finally, residual link with trainable parameters are used in the network to mitigate the influence of weak gradient phenomenon. Furthermore, we demonstrate the effectiveness of PHN in a large number of comparative experiments, and visualize the performance of the model in training process and structure.
	\end{abstract}
	
	\begin{CCSXML}
        <ccs2012>
        <concept>
        <concept_id>10002951.10003317.10003338</concept_id>
        <concept_desc>Information systems~Retrieval models and ranking</concept_desc>
        <concept_significance>300</concept_significance>
        </concept>
        </ccs2012>
	\end{CCSXML}
	
	\ccsdesc[300]{Information systems~Retrieval models and ranking}

	\keywords{Recommendation system, Click-though Rate, Feature Interaction.}
	
	
	\maketitle

\section{Introduction}\label{1}
% CTR点击率预估模型背景

Recommendation system has provided to people a high quality services and brings objective economic benefits to the company. The Click-through rate (CTR) prediction is one of the important basic tasks in recommendation system. By predict the click rate of user, the web or application can sort the candidate item list and push them to target user, so as to provide personalized recommendation service for users. Early CTR prediction models, usually in a simple form, output CTR through Logistic Regression, and use automatic feature engineering methods such as Factorization Machine(FM)\cite{fm} and Gradient Boosting Decision Tree(GBDT)\cite{gbdt} for business implementation. With the development of deep learning, CTR prediction model based on neural network has gradually become the mainstream application model in the real application. The CTR prediction model based deep learning can roughly classify into two categories: one is Wide \& Deep\cite{widedeep} structure, which is analyzed separately based on fixed features. The other is a DIN\cite{din} structure based on the user's historical behavior.

% 主流模型的缺陷
DIN is a model for analyzing user's historical behavior, which predicts CTR by calculating the amount of attention between the target item and the user's history item. In the later development of such models, combined with different analysis structures, including DIEN\cite{dien}, BST\cite{bst} and DIHN\cite{dihn} models. However, they generally relied on the historical behavior of users and needed some other strategies or structures to supplement the problem of cold start.Wide\&deep structure have used parallel structures of different depths to consider both memorization and generalization. In subsequent studies, FNN\cite{fnn}, DeepCrossing\cite{deepcrossing}, DeepFM\cite{deepfm}, DCN\cite{dcn}, xDeepFM\cite{xdeepfm}, DCN-V2\cite{dcnv2}, EDCN\cite{edcn} and other models have similar parallel structure like Wide\&deep, and were utilized to analyze public embedding through different modules. Moreover, the generalization of this structure depended on the effectiveness of parallel structures.

At the end of CTR model, the click rate prediction output have been achieved by linear layer with activation function. During training phase, the activation values between parallel layers tend to fall into the weak gradient interval. This phenomenon will weaken the training effect of each parallel module, and can not improve the generalization while improving the complexity of the model.

% 本文出发点以及针对所做的事
In this paper, we propose a new deep CTR model, named Parallel Heterogeneous Network (PHN). For PHN model, three parallel feature interaction structures were included to analyze CTR features from different perspectives. In order to enhance the independent analysis ability of each parallel module, Soft Select Gating module was constructed after public embedding to enhance the original embedding expression. We also added residual connections with trainable parameters to the model to reduce the weak gradient phenomenon by accumulating gradients during the back propagation process.

%本文主要创新点
This paper mainly contributions are as follow:

	\begin{itemize}
		\item In order to strengthen the expression ability of CTR prediction model, A constructed three different linear feature interaction methods from nonlinear interaction, bite-wise interaction and vector-wise interaction based on parallel structure.
		\item Soft Selection Gating is constructed before the parallel structure, and the features of original embedding are enhanced by self-attention and soft gate structure while retaining the high order crossover characteristics, which improves the ability of the model to express data.
		\item To solve the weak gradient phenomenon in the parallel model, the residual link with trainable parameters are used in the parallel structure to reinforce the model training process.
		\item The effectiveness of Soft Selection Gating and the weak gradient phenomenon are visualized, and the effectiveness of PHN is verified by comparison experiments.
	\end{itemize}

%剩余章节安排
The rest of the contents of this paper are as follow: Part \ref{2} introduce the relevant work of this paper, including the information selection and the feature interaction in CTR model study. Part \ref{3} focus on the overall structure and details of the PHN model. Part \ref{4} introduce the experimental purpose and relevant preparation, and verify the effectiveness of the model through the experimental results. Part \ref{5} is the conclusion.

\section{RELATED WORK}\label{2}%这一部分暂时没改（1.26）
In CTR prediction model study, there are two important parts.

% 信息筛选
%{\bf Information selection:} 在早期CTR模型发展过程中，特征工程作为业务效果提升的主要手段是个性化建模用户兴趣的重点。基于逻辑斯蒂克回归的简单模型可以利用GBDT的叶子节点输出作为二阶特征，有效地提升模型的拟合效果；而基于二阶交叉的线性模型可以利用FM对稀疏的二阶特征进行稠密化映射，来强化模型的训练效果。现如今的CTR同样也较为注重特征工程。基于FM进行特征工程模型有FNN和DeepFM；在PNN中使用特征点乘的方法也可以实现。FiBiNet中的SENET部分也可以看作是特征处理。
{\bf Information selection:} In the early development of CTR model, feature engineering, as the main means to improve business effect, was focused on personalized modeling user interest. And the simple model based on logistic regression was utilized to perform the leaf node output of GBDT\cite{gbdt} as the second-order feature to effectively improve the fitting effect of the model. In a similar fashion, the linear model based on second-order interaction also used FM\cite{fm}, FwFM\cite{fwfm}, and FmFM\cite{fmfm} to map sparse second-order features to enhance the training effect of the model. Morerecently, CTR has also payed more attention to feature engineering. The feature engineering models based on FM include FNN\cite{fnn} and DeepFM\cite{deepfm}. Moreover, the method of using feature point multiplication in PNN\cite{pnn} was also realized. The SENET\cite{senet} part of FiBiNet\cite{fibinet} was regarded as feature processing. The L0-SIGN\cite{l0-sign} uses graph neural network to select information and feature intersection.

% 特征交叉
%{\bf Feature Interaction:} 在推荐系统中，特定的组合往往也可以发挥出高激活值。这里就可以写cross， 点乘交叉、NFM中的价差、FiBiNET中间的交叉、FINT中间的交叉、EDCN中间的交叉。
{\bf Feature Interaction:} Based on the background knowledge of the recommendation system, the combination of features has been used effectively to analyze the preferences of different user groups for specific products, which makes the feature interaction as a key part of CTR prediction model. In a more advanced method such as PNN\cite{pnn}, DCN\cite{dcn} and DCNV2\cite{dcnv2}, NFM\cite{nfm},  FiBiNet\cite{fibinet} and FINT\cite{fint},  xDeepFM\cite{xdeepfm}, and AutoInt\cite{autoint} were separated using for interaction patterns to product layer, cross layer, Hadamard product, field interation, compressed interation network, and Multi-head Self-Attention\cite{msa}, respectively.

\section{PROPOSED METHOD}\label{3}
\begin{figure*}[ht]
  \centering
  \includegraphics[width=\textwidth]{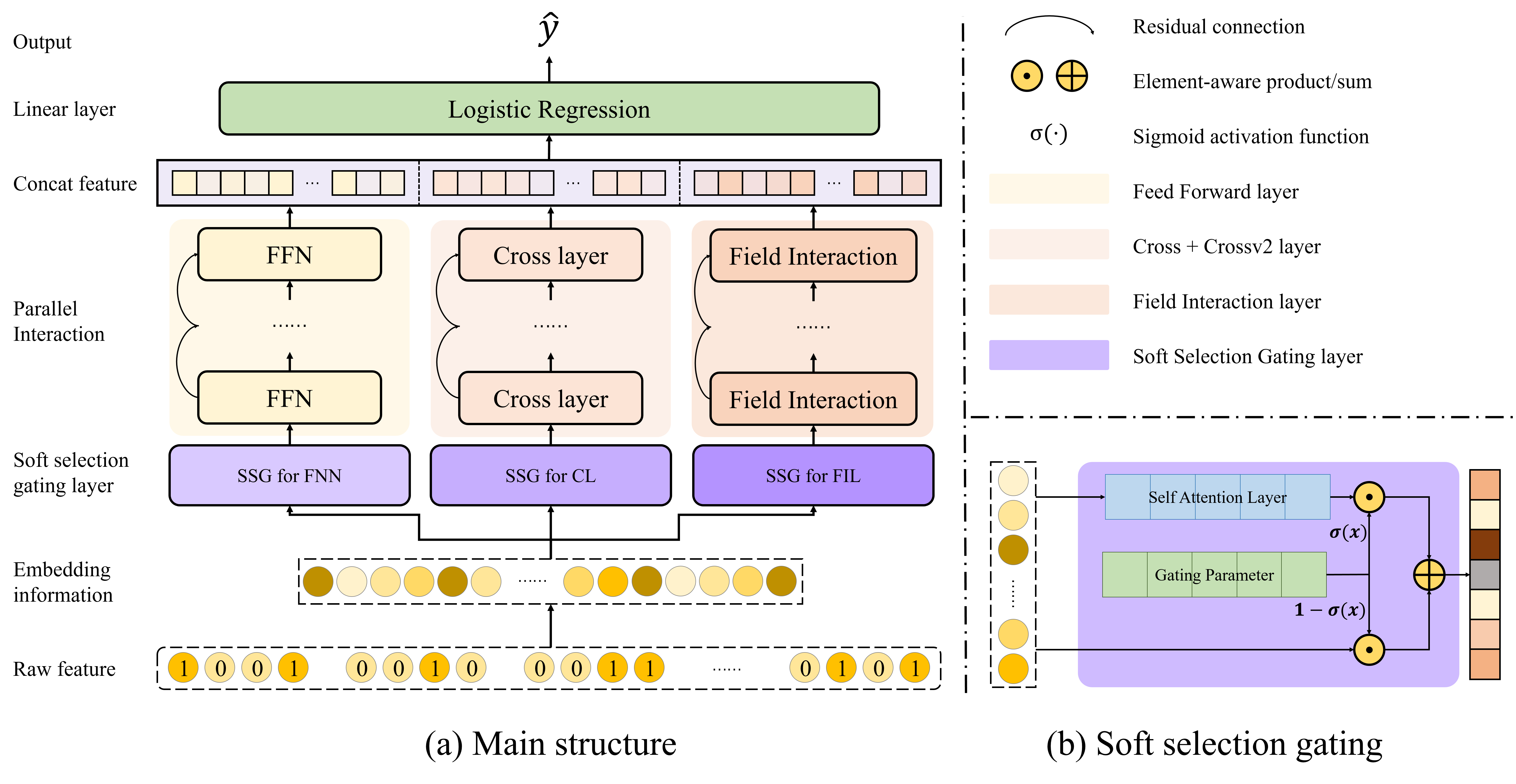}
  \caption{{\bf The overview structure of our proposed PHN model, which consists of Soft Selection Gating (SSG) module and Heterogeneous Interaction Layer(HIL)}}\label{pic:stru}
\end{figure*}

% 主要模型结构
The Parallel Heterogeneous Network (PHN) consists of two main structures. One of them is Soft Selection Gating (SSG) module based on self-attention to enhanced embedding features for different structures, and another one is Heterogeneous Interaction Layer(HIL), using different interaction method to analyze the enhanced features, and finally using Logistic Regression to output the confidence of sample. Fig.\ref{pic:stru} illustrates the main structure and the detail of model.

\subsection{Soft-Gating Information Selection}\label{3.1}
% 基于 transformer 的 soft gating selection 机制
The parallel structure models aim to achieve better generalization by specifying different features interaction structures in a parallel way. This kind of model is similar to bagging model, which means that we can apply more than two structure in CTR model.

The ideal of multiple structure parallelism raise up a new question: whether different structures require different input dense feature. The traditional Multi-head self-attention(MSA) mechanism\cite{msa} used weight based query vector and key vector to aggregate information in a sequence, which was an ideal method to process feature information.

\begin{equation}
Q_E, \;K_E, \;V_E = W_QE, \;W_KE, \;W_VE
\label{eq:msa1}
\end{equation}

\begin{equation}
MSA(Q_E,K_E,V_E) = Softmax(\frac{Q_EK_E}{\sqrt{d_k}})V_E
\label{eq:msa2}
\end{equation}

From the Eq\ref{eq:msa1}\ref{eq:msa2}, MSA has considered different field weights through the second-order intersection of query vector $Q_E$ and key vector $K_E$. However, based on the feature interaction in the CTR prediction model, the direct using by the traditional MSA may over-focus on the feature activation value of the second-order crossover, thus losing the performance of the feature at the higher-order crossover. Inspired by the FRNet\cite{frnet}, an information selecting method named Soft Selection Gating (SSG) is used after the sharing embedding $E\in R^n$ in PHN. This soft-gating information selection is designed for choosing activation between MSA result and sharing raw embedding.

\begin{equation}
E_{sg}=G_{sg}\odot E_{sa} + [I-G_{sg}]\odot E_{se}
\label{eq:ssg}
\end{equation}
where $G_{sg}\in R^n$ is the trainable gating vector, $E_{sa}\in R^n$ is the sharing self-attention embedding, $E_{se}\in R^n$ is the sharing embedding, and $I\in R^n$ is a unit vector. The SSG considers both sharing embedding and self-attention embedding. By using weighting parameters, the model can select the raw feature or the feature enhanced by MSA for different parallel structures. Subsequent experiments will discuss whether to share the weight of self-attention and the gating mechanism, and confirm the effectiveness of SSG.

\subsection{Heterogeneous Interaction Layer}\label{3.2}
% 使用三种结构并行构建的特征异构分析层
In PHN, the parallel use three kinds of interaction layers to implementation the parallel structure: 1) cross layer is the basic part of DCN\cite{dcn} and DCNV2\cite{dcnv2}, which focuses on the bit-aware feature interaction; 2) field interaction layer is the basic part of FINT\cite{fint}, which focuses on the vector-aware field interaction; 3) feed forward layer is used to fitting the non-polynomial information.

\subsubsection{Cross Interaction Layer}\label{3.2.1}
Feature interaction is a main key point in study of mainly CTR prediction model. As a previous study, the DCN\cite{dcn} and the DCNV2\cite{dcnv2} proposed two kinds of explicit interaction methods, which achieved the data mode of high-order interaction by realizing the intersection of multi-layer hidden features and original features.

\begin{equation}
y_{dcnv2} = x_0 \odot (W \times x_i + b) + x_i
\label{eq:cross1}
\end{equation}

\begin{equation}
y_{dcn} = x_0 \odot x_i^T * w + x_i + b
\label{eq:cross2}
\end{equation}
where, $x_0$ is the input feature of the first cross layer; $x_i$ is the output feature of the i-th cross layer, $W$ and $w$ represent trainable parameter vectors and matrices; $\odot$ means Hadamard product and $\times$ means matrix multiplication. In Eq.\ref{eq:cross} the DCN and DCNV2 used different parameter forms to interact features, but in general, it achieves bit-aware feature interaction. The PHN combines the formulas of DCN and DCNV2 to construct the bit-aware interaction module.

%x_{i+1} = x_0 \odot (W \times x_i + b) + x_0 \times x_i \times w + b
\begin{figure}[ht]
  \centering
  \includegraphics[width=0.45\textwidth]{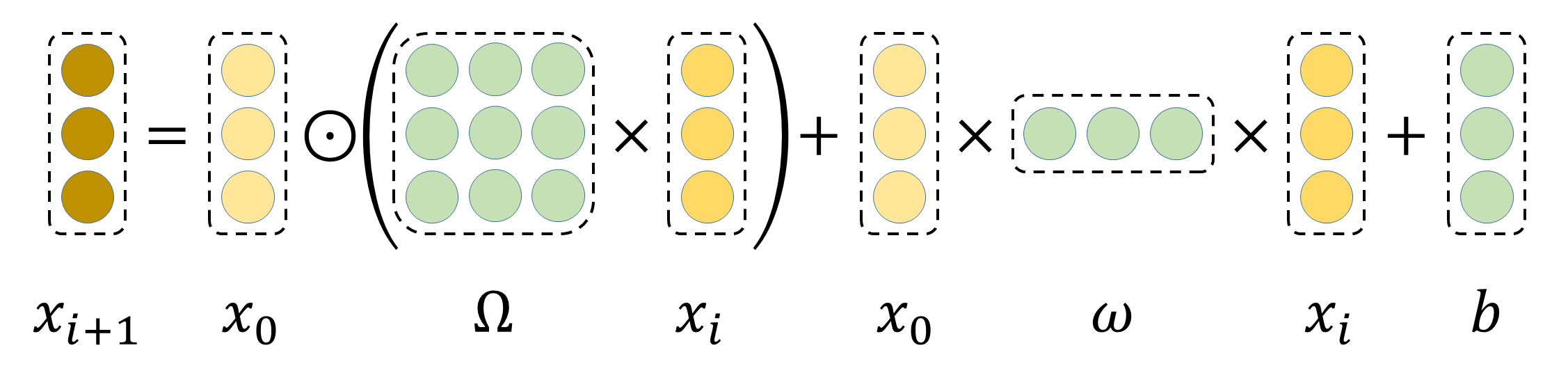}
  \caption{{\bf The calculation diagram of cross layers in PHN}}\label{pic:cross}
\end{figure}

As mentioned in Fig.\ref{pic:cross}, we use the parameter part of the two crossover model and bias to construct cross layer in PHN.

\subsubsection{Field Interaction Layer}\label{3.2.2}
Besides bit-aware interaction, vector-aware interaction is also a key part of the model construction. Field Interaction module is mentioned in FiBiNet\cite{fibinet} and FINT\cite{fint}, which using the cross method to implement the vector-aware interaction. PHN use the Field Interaction layer in FRNet as a parallel part to enhance the  generalization effect of the whole network on the feature crossing pattern.

% V_{i+1} = V_i \odot (W \times V_0) + U \odot V_i
\begin{figure}[ht]
  \centering
  \includegraphics[width=0.45\textwidth]{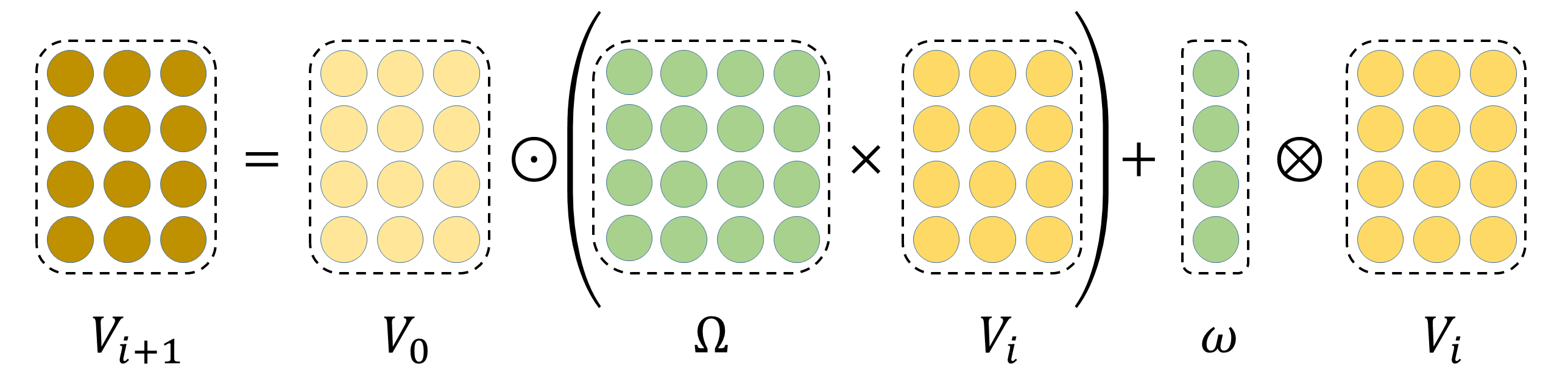}
  \caption{{\bf The calculation diagram of field interaction layers in PHN}}\label{pic:fi}
\end{figure}

As shown in Fig.\ref{pic:fi}, field interaction layer uses the residual link with trainable parameter vector, which product on different fields feature to screen the output features of the upper layer. In the subsequent experiments, we will discuss and experiment residual link forms in all parallel layers.

\subsubsection{Feed Forward Layer}\label{3.2.3}
The third part of parallel network is composed of Feed Forward Network. By alternating linear and nonlinear analysis of the original features, FFN complements the analysis of the previous two crossover modes to improve the overall network generalization function.

\begin{equation}
x_{i+1} = \sigma(\omega x_i+b)
\label{eq:ffn}
\end{equation}
where, $\sigma$ is the activation function, which is LeakyReLU in PHN.

\subsection{Weak Gradient Problem}\label{3.3}
% 最后的全连接层利用batchnorm来解决网络bagging所带来的梯度衰弱现象。
Basic on the prediction task definition, the key point of improving AUC value is increasing the confidence of positive label samples and decreasing the confidence of negative label samples, which makes model more robust. In the last stage of CTR prediction model, the traditional model usually constructs confidence coefficient of click by using Sigmoid activation function.

\begin{equation}
\frac{\partial \sigma}{\partial x}=\frac{e^{-x}}{(1+e^{-x})^2}=\sigma(1-\sigma)
\label{eq:sigmoid1}
\end{equation}

\begin{equation}
\sigma(z)=\frac{1}{1+e^{-z}}\;\;0<\sigma<1
\label{eq:sigmoid2}
\end{equation}

As shown in Eq.\ref{eq:sigmoid}, as the absolute value of the final linear layer output increases, the gradient propagated back based on the activation function also decreases.

\begin{figure}[ht]
  \centering
  \includegraphics[width=0.45\textwidth]{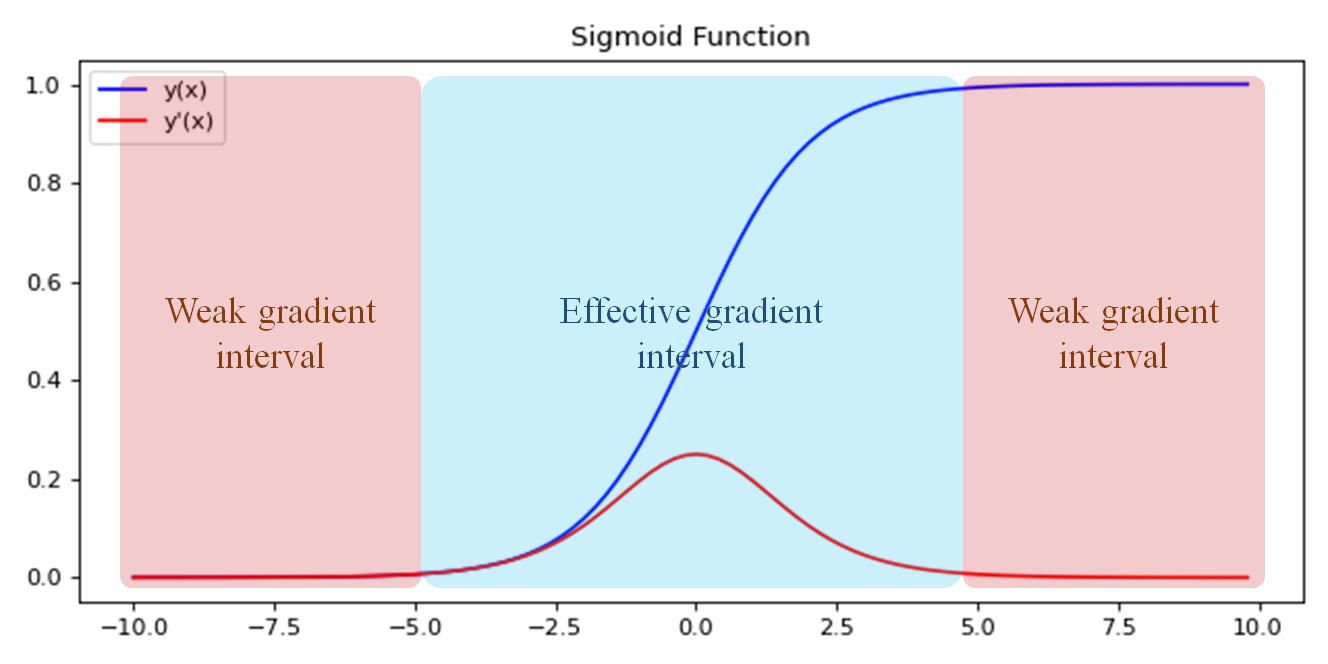}
  \caption{{\bf Gradients of different interval in Sigmoid}}\label{pic:sigmoid}
\end{figure}

The Fig.\ref{pic:sigmoid} shows that, we can think of the entire Sigmoid function as two interval, the effective gradient interval (bule) with a normal gradient, and the Weak gradient interval (red) with a gradient approaching zero. When the output value of the parallel model is accumulated at the last linear layer, it is easy to make the samples originally in the effective gradient interval to fall into the Weak gradient interval, thus weakening the learning of each part for valid samples. In this paper, this phenomenon is referred to weak gradient in parallel structures.

To mitigate this phenomenon and achieve effective training, the PHN using residual link in each substructure, enhancing the gradient accumulation of each parallel structure in the process of back propagation. To further accommodate this phenomenon, we also tried to add gating parameters to residual links and used batch normalization of different modes in the final linear layer. We will discuss this further in Part\ref{4}.

\section{EXPERIMENTS}\label{4}
In this section, we evaluate PHN on two benchmark data sets. We aim to answer the following research questions:

	\begin{itemize}
		\item {\bf RQ1:} Will parallel structure-based PHNS perform better than previous CTR prediction models over different classical data sets?
		\item {\bf RQ2:} Is the parallel structure actually caused the weak gradient phenomenon, and the problem is effectively alleviated by residual connection or batch normalization?
		\item {\bf RQ3:} Under what circumstances can the information selecting module based on soft gating machine reasonably enhance the function of feature representation.
	\end{itemize}

\subsection{Datasets Description.}\label{4.1}
To evaluate the effectiveness of the model in this paper, two benchmark off-line datasets are selected for experiment: Criteo\footnote{https://www.kaggle.com/c/criteo-display-ad-challenge} and Avazu\footnote{https://www.kaggle.com/c/avazu-ctr-prediction}. Detailed information on the two benchmark datasets is shown in Table\ref{table:data}.

\begin{table}[h]
  \centering
  \caption{Statistics of the benchmark datasets}\label{table:data}
    \begin{tabular}{lcccc}
    \hline
    Dataset & Sample size & Fields & Features & Positive Ratio \\
    \hline
    \hline
    Criteo & 45,840,618 & 39 & 1,086,810 & 25.6\% \\
    Avazu  & 40,428,966 & 23 & 1,544,257 & 16.9\% \\
    \hline
    \end{tabular}
\end{table}

For the CTR prediction task, two classical evaluation metrics such as: Logloss and AUC, were used to verify the effective generalization and robustness of the suggested model.

\subsection{Compared Models.}\label{4.2}
To verify the effectiveness of the proposed PHN model, we compare it with linear model (LR, FM\cite{fm}, FwFM\cite{fwfm}, FmFM\cite{fmfm}), deep model (DNN, W\&D\cite{widedeep}, DeepFM\cite{deepfm}, xDeepFM\cite{xdeepfm}, AutoInt\cite{autoint}), and interaction model (DCN\cite{dcn}, DCNV2\cite{dcnv2}, FiBiNet\cite{fibinet}, FINT\cite{fint}) on CTR task.

%\begin{itemize}
%		\item \textbf{LR}, as the basic regression method, used as a benchmark experiment for other models.
%		\item \textbf{FM} uses the interaction vector densification method to improve the training effectiveness of the second order interaction term.
%		\item \textbf{FwFM} improves the interaction mode of embedding on the basis of FM.
%		\item \textbf{FmFM} improves the weight vector of embedding interaction mode based on FwFM.
%		\item \textbf{DNN} is the most basic deep learning method to perform CTR prediction task.
%		\item \textbf{W\&D} uses linear layer and DNN to realize wide side and deep side of the model respectively.
%		\item \textbf{DCN} changed the wide side of W\&D to cross network, and explicitly carried out multi-order interaction of features.
%		\item \textbf{DCNV2} improved the interaction mode of cross network and further enhanced the interaction dimension of the network based on DCN.
%		\item \textbf{DeepFM} uses FM to improve the linear layer of W\&D and strengthen its generalization ability
%		\item \textbf{xDeepFM} introduced CIN network on the basis of DeepFM to strengthen the generalization ability of the model.
%		\item \textbf{AutoInt} uses the multi-headed self-attention structure in the generalization model distribution.
%		\item \textbf{FiBiNet} introduces SENET to initialize the embedding.
%		\item \textbf{FINT} uses weight matrix and residual link with parameters to interaction field information in multiple layers.
%	\end{itemize}

\subsection{Experimental Setting.}\label{4.3}
In the rest of this section, there are three sub-sections to evaluate the validity of the proposed model, the weak gradient problem solution, and the Soft Selection Gating module, while the experimental model designs are different. In section\ref{4.4}, the same number of cross layers were used to compare the performance of PHN and other CTR model on the benchmark data set, and grid search was performed on the number of cross layers. In section\ref{4.5}, we first explored the effectiveness of RL and BN on PHN through comparative experiments. Then, different parallel parts of the PHN and different model patterns were trained an epoch independently, and 200 samples from the model were extracted for visualization of the results. In section\ref{4.6}, the two benchmark models verify the validity of SSG in different modes and visually display the calculation results of SSG. All experiments were carried out on FuxiCTR\cite{fuxictr} open source framework.

\subsection{Performance Comparison (RQ1)}\label{4.4}
% 与现有方法的对比实验，以及单独训练和并行训练的激活值影响
\subsubsection{Effectiveness of PHN.}\label{4.4.1}
To verify the validity of the model, we followed the original structure of each comparison model, controlled the cross layer of all models in three layers, and recorded the validation and testing results of each model on two benchmark datasets. The specific experimental results were reported in Table\ref{table:exp1}.

\begin{table}[h]
  \centering
  \small
  \fontsize{9}{12}\selectfont
  \caption{Experiment result of different CTR prediction models on Criteo and Avazu}\label{table:exp1}
    \begin{tabular}{c|cc|cc}
    \hline
    \multirow{2}*{model} & \multicolumn{2}{c|}{Criteo} & \multicolumn{2}{c}{Avazu} \\
    \cline{2-5}
        & logloss & AUC & logloss & AUC  \\
    \hline
    LR   & 0.457334 & 0.792831 & 0.382039 & 0.777148 \\
    FM   & 0.450260 & 0.801086 & 0.378750 & 0.782448 \\
    FwFM & 0.442566 & 0.809314 & 0.373862 & 0.790315 \\
    FmFM & 0.444253 & 0.807395 & 0.376521 & 0.785998 \\
    \hline
    DNN     & 0.442271 & 0.809547 & 0.372686 & 0.792553 \\
    W\&D    & 0.442627 & 0.809133 & 0.372663 & 0.792079 \\
    DCN     & 0.442382 & 0.809390 & 0.372884 & 0.791767 \\
    DCNV2   & 0.440825 & 0.811139 & 0.372511 & 0.792352 \\
    DeepFM  & 0.444391 & 0.807686 & 0.372202 & 0.792856 \\
    xDeepFM & 0.444541 & 0.807728 & 0.373387 & 0.791503 \\
    AutoInt & 0.442502 & 0.809237 & 0.372830 & 0.791918 \\
    FiBiNET & 0.442335 & 0.809809 & 0.371139 & 0.794850 \\
    FINT    & 0.442471 & 0.808409 & 0.372808 & 0.792043 \\
    \hline
    PHN(ours) & {\bf 0.439927} & {\bf 0.812039} & {\bf 0.370481} & {\bf 0.795964} \\
    \hline
    \end{tabular}
\end{table}

The experiment shows that, PHN achieved a good performance in both validation and testing experiments with two benchmark large datasets.

\subsubsection{Grid Search.}\label{4.4.2}

\begin{figure}[ht]
\setlength{}
  \centering
  \addtocounter{subfigure}{0}\subfigure[Grid search experiment on Criteo]{
    \includegraphics[width=.45\textwidth]{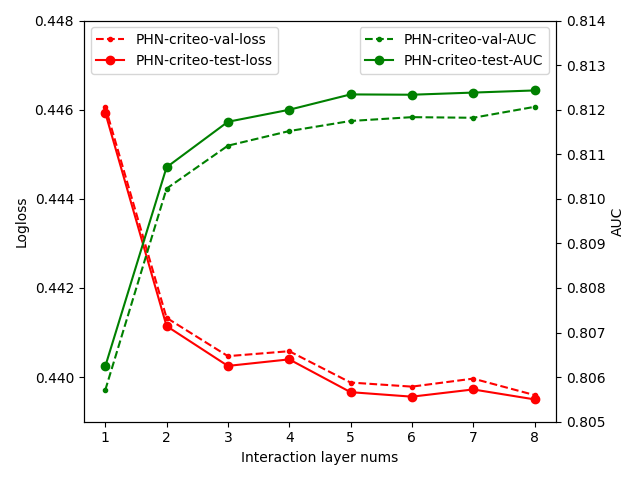}
  }
  \addtocounter{subfigure}{0}\subfigure[Grid search experiment on Avazu]{
    \includegraphics[width=.45\textwidth]{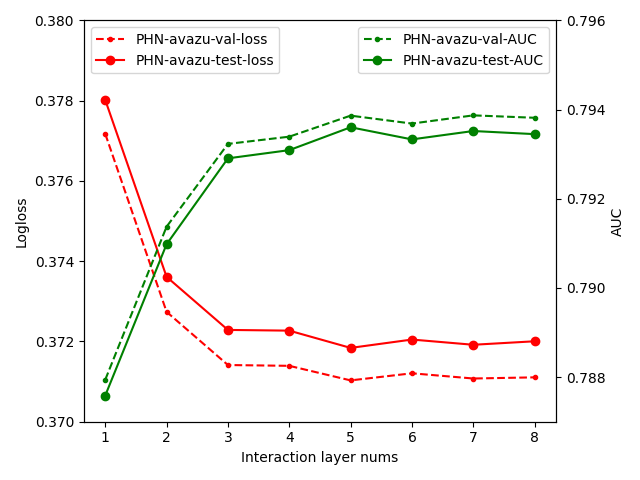}
  }
  \caption{{\bf The grid search performance of PHN with different interaction layers on two benchmark datasets.}}
  \label{pic:grid}
\end{figure}

Fig\ref{pic:grid} shows the grid search experiment result of PHN. With the increase of the number of cross layers, the robustness of the model also increases, which may benefit from the improvement of the expression ability of cross layers for higher-order crosses. However, as the number of layers increases, the model complexity also increase, which slow down the training process of the model. The values of AUC and Logloss shown in Fig\ref{pic:grid} tend to be stable when the number of cross layers is 5, so we can consider the optimal number of cross layers for PHN to be 5.

\subsection{Weak Gradient Phenomenon (RQ2)}\label{4.5}

\subsubsection{Efficiency analysis}\label{4.5.1}
\begin{figure}[ht]
\setlength{}
  \centering
  \addtocounter{subfigure}{0}\subfigure[PHN without BN]{
    \includegraphics[width=.13\textwidth]{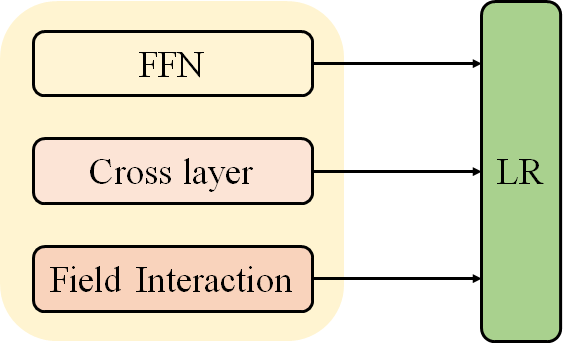}
  }
  \addtocounter{subfigure}{0}\subfigure[PHN with public BN]{
    \includegraphics[width=.13\textwidth]{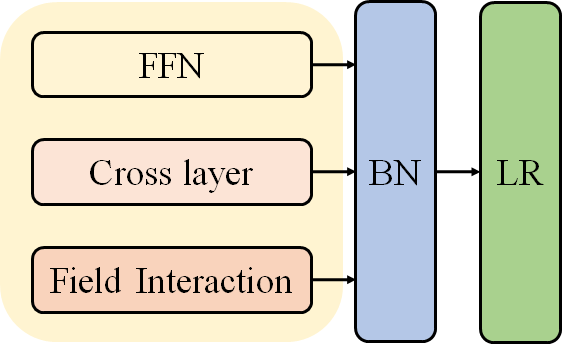}
  }
  \addtocounter{subfigure}{0}\subfigure[PHN with private BN]{
    \includegraphics[width=.13\textwidth]{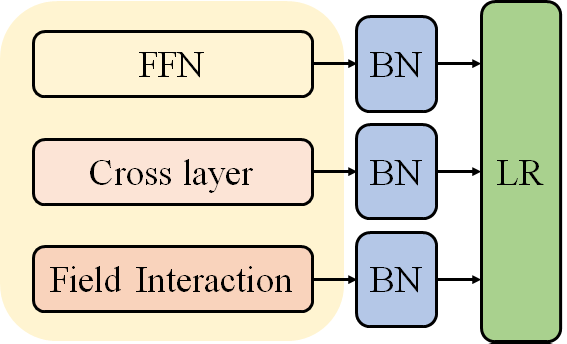}
  }
  \caption{{\bf The details of batch normalization in PHN}}
  \label{pic:phn_bn}
\end{figure}

To reduce the impact of weak gradient phenomenon, there has been an attempt in PHN to enhance the data gradient flow in training through residual links or batch normalization to reduce the training pressure of each parallel part. In experiments, we tried to introduce gating parameters for residual links and discussed the independence of batch normalization in different parallel modules. The Fig\ref{pic:phn_bn} shows the batch normalization in different modes.

This part of the comparison test based on different solutions, gives the performance of the algorithm in different cases. Table\ref{table:exp2} shows the result of comparison experiments.

\begin{table}
  \centering
  \caption{Experiment result of different solutions to the weak gradient problem on Criteo and Avazu}\label{table:exp2}
    \begin{tabular}{|c|cc|cc|}
    \hline
    \multirow{2}*{Model} & \multicolumn{2}{c|}{Criteo} & \multicolumn{2}{c|}{Avazu} \\
    \cline{2-5}
      & logloss & AUC & logloss & AUC \\
    \hline
     \makecell[l]{$PHN_{base}$} & 0.440034 & 0.811914 & 0.372209 & 0.793206 \\
     \makecell[l]{$PHN_{rl}$}   & 0.439763 & 0.812111 & 0.372294 & 0.793044 \\
     \makecell[l]{$PHN_{prl}$}  & {\bf 0.439540} & {\bf 0.812428} & 0.372084 & 0.793392 \\
    \hline
    \hline
     \makecell[l]{$PHN_{base+bn}$} & 0.440111 & 0.811879 & {\bf 0.371117} & {\bf 0.795087} \\
     \makecell[l]{$PHN_{rl+bn}$}   & 0.441548 & 0.810359 & 0.372410 & 0.793084 \\
     \makecell[l]{$PHN_{prl+bn}$}  & 0.440307 & 0.811711 & 0.371189 & 0.794911 \\
    \hline
    \hline
     \makecell[l]{$PHN_{base+pbn}$} & 0.443966 & 0.811865 & 0.373950 & 0.794839 \\
     \makecell[l]{$PHN_{rl+pbn}$}   & 0.445333 & 0.809268 & 0.379022 & 0.791024 \\
     \makecell[l]{$PHN_{prl+pbn}$}  & 0.444590 & 0.811813 & 0.377631 & 0.794621 \\
    \hline
    \end{tabular}
\end{table}

% 残差链接可以在反向传导过程中通过梯度累积的方式来weaken弱梯度问题，而带有参数的情况下，整体网络可以更好的在前馈和反向的过程中拟合数据流，强化不同并行结构的拟合效果。但是在后续加上batchnormalization并没有起到很好的效果，这可能是由于不同并行结构中的数据流分布不一，而使用normalization进行强行统一后反而削弱了数据的表达形式。这也一定程度上说明了bn层与线性层不同，是不可拆分的。最后两组实验也说明了，残差链接强化不同并行结构时所拟合的数据流特异性被强化，而bn反而起到了一定的副作用，因此在PHN中，残差和bn最好是在独立的并行结构中实现。
Different model subscripts represent different structures. $base$ means the basic model, $rl$ means normal residual links is added to the PHN, $Prl$ means parameter residual links is added to the PHN, $bn$ means batch normalization is followed before the final full connection layer, and $pbn$ means private batch normalization.

The experiments show that, In the reverse transmission process, the weak gradient problem can be improved by using gradient accumulation in the residual link. In the case of parameters, the overall network can better fit the data flow in the feed forward and reverse process, and strengthen the fitting effect of different parallel structures. However, the subsequent addition of Batch normalization has not been very effective. This may be due to the uneven distribution of data flows in different parallel structures, but forced unification with normalization weakens the representation of data. This also explains to some extent that batch normalization layer is not separable from linear layer. The last two groups of experiments also showed that the specificity of the data stream fitted was enhanced when the residual link strengthened different parallel structures, while BN had certain side effects. Therefore, in PHN, the residual and Batch Normalization had better be realized in an independent parallel structure.

\subsubsection{Visualization of activation value}\label{4.5.2}
\begin{figure}[ht]
\setlength{}
  \centering
  \addtocounter{subfigure}{0}\subfigure[Independent parallel structure]{
    \includegraphics[width=.3\textwidth]{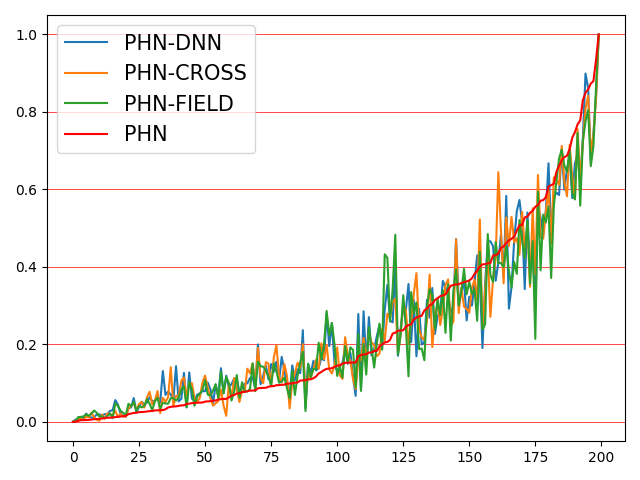}
  }
  \addtocounter{subfigure}{0}\subfigure[Add results of independent parallel structures]{
    \includegraphics[width=.3\textwidth]{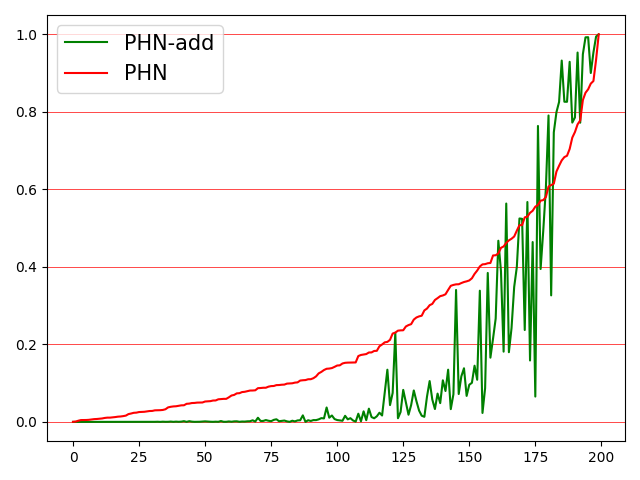}
  }
  \addtocounter{subfigure}{0}\subfigure[PHN with a residual link]{
    \includegraphics[width=.3\textwidth]{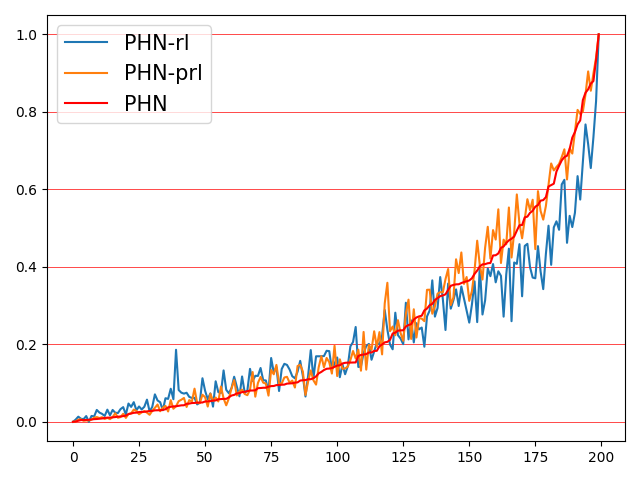}
  }
  \caption{{\bf Experiments on the last activation value of PHN.}}
  \label{pic:phn_vis}
\end{figure}

% 单独训练和并行训练的激活值影响(拿一个epoch的不同样本进行梯度计算，然后将最后的激活值进行可视化)(利用图可视化，将200个样本按照大模型的置信度进行排序，然后再输入分别的以及分别的单纯累加的。用曲线说明，累加的曲线应该在大模型的上面。要做一个没经过处理的版本和处理后的版本。)
A more robust model should output more closely to the confidence of the label worthiness. Fig.\ref{pic:phn_vis} shows the confidence curve of PHN in 200 samples after training one epoch in different configurations.

The red line in Fig.\ref{pic:phn_vis} represents the PHN model without residual link and batch normalization. Fig.\ref{pic:phn_vis}(a) shows that, the single crossover structure showed higher negative confidence and lower positive confidence than PHN. This shows that PHN is superior to partial cross structure in sample resolution. Fig.\ref{pic:phn_vis}(b) shows that, the sigmoid calculation after summing up the activation values of the parallel models can show more robustness than PHN. This means that the fitting effect of a single PHN model on the data set is weakened by weak gradient phenomenon.  Fig.\ref{pic:phn_vis}(c) shows that, residual connection can enhance the high confidence of negative samples, but also reduce the confidence of positive samples, and the residual connection with parameters can effectively improve the performance of the model on the PHN infrastructure.

\subsection{Selection Information (RQ3)}\label{4.6}
% 信息选择层的有效性
\subsubsection{Data skew visualization.}\label{4.6.1}
% 将输入的数据进行倾斜后进行可视化，然后展示其特征交叉的热力图（没倾斜的交叉和倾斜的交叉应该差不多，然后只用trans的交叉激活值应该比较小）

\begin{figure*}[ht]
\setlength{}
  \centering
  \addtocounter{subfigure}{0}\subfigure[Scaling ratio of FFN layer]{
    \includegraphics[width=.3\textwidth]{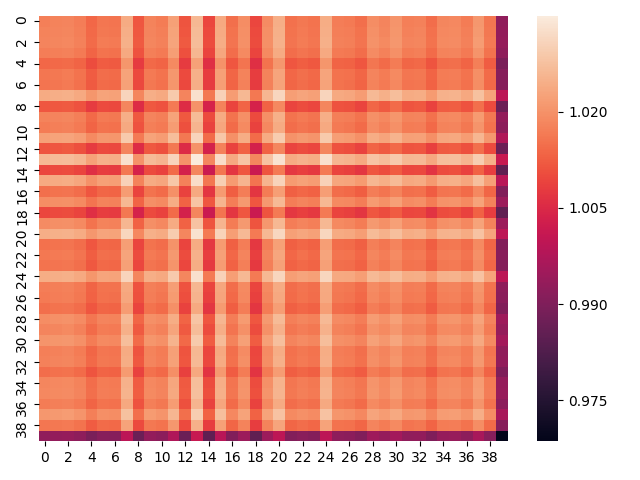}
  }
  \addtocounter{subfigure}{0}\subfigure[Scaling ratio of Cross layer]{
    \includegraphics[width=.3\textwidth]{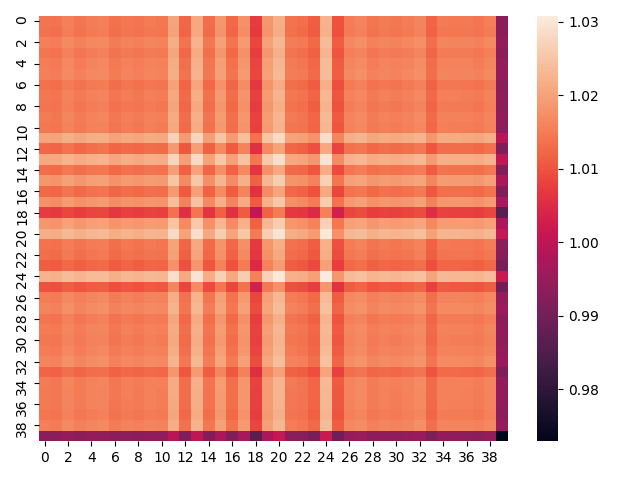}
  }
  \addtocounter{subfigure}{0}\subfigure[Scaling ratio of Field interaction layer]{
    \includegraphics[width=.3\textwidth]{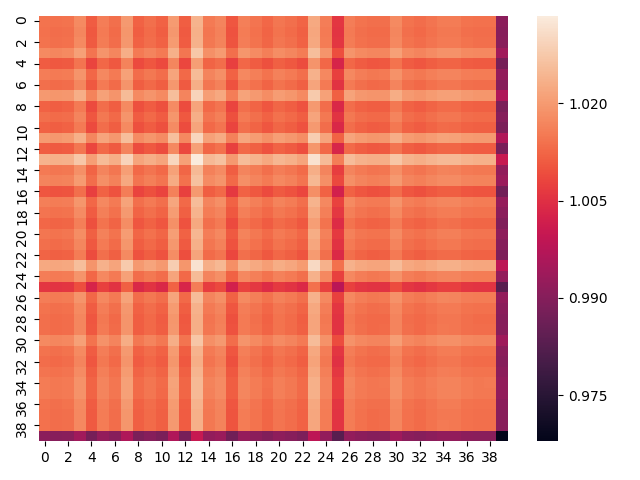}
  }
  \caption{{\bf The heatmap of different parallel layers input feature scaling ratio}}
  \label{pic:phn_heatmap}
\end{figure*}
Based on the trained PHN structure, we visualized the tensor amplification ratio after SSG output. As shown in Figure\ref{pic:phn_heatmap}, the characteristics of the three cross-layers have some similarity, such as the high proportion of field 13 and the low proportion of field 39. At the same time, the feature scaling of the three parts is somewhat different, as in field 8 and field 24.

\subsubsection{Selection pattern.}\label{4.6.2}
% 直接探讨不同交叉模式下最后的结果（embedding是否公用，是否直接用trans，trans是否公用，gating是否公用）(在这里为了表现初期交叉层的影响，在实验时取消Pnn层)
The SSG module is designed for enhance the representation of embedding feature, which select the feature from raw embedding and self-attention embedding. Depending on the design, the selection pattern of self-attention layer and gating layer in this module can be classified as public or private. To further validate the effectiveness of SSG, we also conducted a comparison experiment under a pure parallel structure. Different subscripts represent different selection patterns: "embed" means using public embedding feature; "sa" means using public self-attention feature based on raw embedding; "sg" means using public soft gating to enhance the self-attention feature; the subscripts with a prefix "P" means that the PHN contains private layers for each parallel layer. Table \ref{table:ssg} shows the results of comparison experiment.

\begin{table}[h]
  \centering
  \caption{Experiment result of different schemes of information selection module on Criteo and Avazu}\label{table:ssg}
    \begin{tabular}{|c|cc|cc|}
    \hline
    \multirow{2}*{Model} & \multicolumn{2}{c|}{Criteo} & \multicolumn{2}{c|}{Avazu} \\
    \cline{2-5}
      & logloss & AUC & logloss & AUC \\
    \hline
     \makecell[l]{$PHN_{embed}$}    & 0.440543 & 0.811647 & 0.371538 & 0.794388 \\
    \hline
    \hline
     \makecell[l]{$PHN_{sa}$}    & 0.441301 & 0.810525 & 0.373018 & 0.792504 \\
     \makecell[l]{$PHN_{Psa}$}   & 0.441445 & 0.810554 & 0.372369 & 0.792640 \\
    \hline
    \hline
     \makecell[l]{$PHN_{sa+sg}$}   & 0.440210 & 0.811782 & 0.371608 & {\bf 0.794622} \\
     \makecell[l]{$PHN_{Psa+sg}$}  & {\bf 0.440031} & {\bf 0.811902} & {\bf 0.371351} & 0.794570 \\
     \makecell[l]{$PHN_{sa+Psg}$}  & 0.440256 & 0.811771 & 0.371398 & 0.794385 \\
     \makecell[l]{$PHN_{Psa+Psg}$} & 0.440692 & 0.811595 & 0.371405 & 0.794458 \\
    \hline
    \end{tabular}
\end{table}

The experiments show that, single self-attention layer (public of private) cannot replace the embedding feature represent, but it can help to enhance the feature by using the soft selection gating, and the AUC value of the algorithm increases by 0.123\% on average. From a theoretical point of view, a feature without a high activation value in the first-order feature cannot be completely transferred, because it may show a high activation value in the high-order interaction with other features.

\section{Conclusion}\label{5}
% 总结
In this paper, we described the parallel structure of the current mainstream CTR model and the weak gradient phenomenon that must be faced in the parallel structure, and introduce a parallel structure model named Parallel Heterogeneous Network (PHN) in response to these phenomena. PHN model used Soft Selecting Gating (SSG) structure to isomerize features, and used Feed Forward network, cross interaction layers and field interaction layers to build the subsequent parallel part. The performance experiment results show that PHN shows the State of the Art on two large benchmark data sets, and explores the interaction layer num of the model. The comparative experimental results show that SSG can effectively improve the representation based on public embedding, and the residual link with trainable parameters can improve the representation ability of the model while maintaining the robustness of the results. Based on the overall experimental results, this work brings us to ont step closer to being able to determine the optimal structure of PHN.

\bibliographystyle{ACM-Reference-Format}
\bibliography{PHN}
	
\end{document}